\documentclass[11pt]{article}

\usepackage{comment}
\usepackage{algorithm}
\usepackage{algorithmic}
\usepackage{caption}
\usepackage{subcaption, booktabs}
\usepackage{amssymb}
\usepackage{authblk}
\usepackage{hyperref}
\usepackage{pifont}
\newcommand{\cmark}{\ding{51}}%
\newcommand{\xmark}{\ding{55}}%
\usepackage{graphicx}
\usepackage{url}
\usepackage{amsmath}
\usepackage{cleveref}
\usepackage{siunitx}
\usepackage{dcolumn}
\newcolumntype{d}[1]{D{.}{.}{#1}}

\title{\vspace{-10ex}A Robust Classification Framework for \\Byzantine-Resilient Stochastic Gradient Descent}
\date{\vspace{-5ex}}
\author{Shashank Reddy Chirra}
\author{Kalyan Varma Nadimpalli\footnote{Corresponding author; \url{kalyan.varma@iiitb.ac.in}}}
\author{Shrisha Rao}
\affil{International Institute of Information Technology Bangalore, Bangalore, India}
         
\newcommand{\BibTeX}{\rm B\kern-.05em{\sc i\kern-.025em b}\kern-.08em\TeX}

\begin{document}
\maketitle


\begin{abstract}
This paper proposes a Robust Gradient Classification Framework (RGCF) for Byzantine fault tolerance
in distributed stochastic gradient descent.  The framework consists of a pattern recognition filter which we
train to be able to classify individual gradients
as Byzantine by using their direction alone.
This filter is robust to an arbitrary
number of Byzantine workers for convex as well
as non-convex optimisation settings, which is a
significant improvement on the prior work that is
robust to Byzantine faults only when up to 50\% of
the workers are Byzantine. This solution does not
require an estimate of the number of Byzantine
workers; its running time is not dependent on the
number of workers and can scale up to training
instances with a large number of workers without
a loss in performance. We validate our solution
by training convolutional neural networks on the
MNIST dataset in the presence of
Byzantine workers. The corresponding code can be found at: \href{https://github.com/nkalyanv/RGCF}{https://github.com/nkalyanv/RGCF}
\end{abstract}


\section{Introduction}
In recent years, there has been a surge in popularity for using deep learning algorithms to solve complex problems in a multitude of domains including computer vision, natural language processing, and speech recognition. Solving these tasks has seen an increase in demand for larger and deeper networks; for example, the GPT-3 \cite{NEURIPS2020_1457c0d6} model uses around 175 billion parameters to produce human-like text. Consequently, training such large models requires enormous datasets, which are often populated by data generated by devices located at different geo-locations across the globe. This makes it practically infeasible to aggregate the data onto a single server. Training is then done over a cluster of machines with the computational overhead being split across them and parallelism being used to reduce training time.  This, however, adds computational as well as communication overheads to pass and aggregate information across these machines. 

\begin{figure}
    \centering
    \includegraphics[scale=0.5]{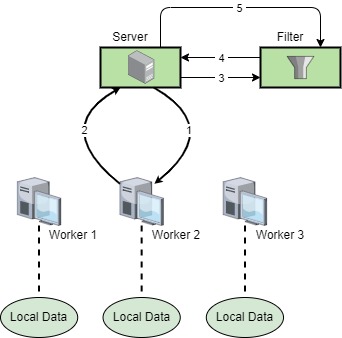}
    \caption{Robust Gradient Classification Framework}
    \label{architectureFig}
\end{figure}

 The standard framework for distributed training is that of a parameter-server architecture where there exists a single central server and several worker nodes \cite{li2014scaling}. The central server stores the model parameters and routinely distributes these parameters to the worker nodes.  The training is largely done on the worker nodes and the generated training information is passed back to the central server that updates its parameters accordingly. While most methods are based on this key central concept, they differ in their specific implementations to tackle various use cases. One popular example is federated learning \cite{mcmahan2017communication} which aims to ensure the privacy of the worker nodes by using aggregation rules that do not reveal the information passed by individual workers.

Such a framework must be robust against Byzantine failures \cite{lamport1982byzantine}. Here, `Byzantine' is an umbrella term for the most unrestricted class of failures.  A Byzantine failure can occur due to any means, and act in any manner, to prevent the convergence of the parameter server.  This can be due to an error in transmission, computational errors, biased data sampled by the worker, and in the extreme case, because of malicious information passed by workers that possibly collude to harm the convergence of the parameter server. 

There have been multiple solutions proposed for Byzantine fault tolerance for the parameter server architecture in the past ~\cite{blanchard2017machine, guerraoui2018hidden, yin2018byzantine, yang2019byzantine}, and also federated learning setups \cite{data2021byzantine, so2020byzantine}.  The majority of these algorithms require an upper bound $f$ on the fraction of workers that can be Byzantine.  This $f$ is used to filter potential Byzantine gradients.  As a result, they face a slowdown that is proportional to $f$.

These algorithms share a common structure: initially, gradients are collected from all $n$ workers (or in some cases, a subset).  An aggregation rule is then applied to estimate the true gradient.  The robustness of this rule is dependent on the correct estimate of the fraction $f$ of Byzantine workers.  Also, then the speed of convergence of the parameter server is heavily dependent on this fraction $f$, as is explained in the following section. 

We introduce a novel Robust Gradient Classification Framework (RGCF) that takes a different approach to Byzantine fault tolerance in distributed training frameworks.  RGCF uses a pattern recognition filter to classify whether a single gradient is Byzantine.  Gradients classified as Byzantine by the filter are dropped and are not passed to the central server.  The filter is trained in a supervised learning using several iterations on a local clean dataset.  The filter then generalises this knowledge to classify gradients obtained from unseen data. 

The major contributions of this paper are:
\begin{itemize}
    \item \emph{Robustness to an arbitrary number of Byzantine workers:} RGCF is robust to \textit{any} number of Byzantine workers, not just in cases where a majority of the workers are non-Byzantine (in practical settings, an arbitrary number of workers may be Byzantine). RGCF can classify all gradients as Byzantine even without the presence of a \textit{single} honest worker.  RGCF is usable in \textit{convex} as well as \textit{non-convex} optimisation settings.
    
    \item \emph{No dependence on pre-defined $f$:} In a practical setting $f$ may be unknown, hence it can be hard to decide on an ideal value of $f$.  RGCF does not use such a bound $f$ in any manner.
    
    \item \emph{Lower computational complexity:}
    RGCF is based on a filter that can classify the nature of a single gradient, hence at each iteration, one gradient is obtained from one selected worker, hence there a communication overhead of $\mathcal{O}(d)$.  Other approaches to robustness~\cite{allen2020byzantine, blanchard2017machine, guerraoui2018hidden} are based on aggregates of gradients from each worker, hence at each iteration one gradient is sampled from each worker, so that there is a communication overhead of $\mathcal{O}(nd)$, where $n$ denotes the number of workers and $d$ is the number of parameters of the server.
    
\end{itemize}

The rest of the paper is as follows: Section~\ref{RecentWork} summarises some of the recent approaches to tackle the problem of Byzantine fault tolerance in parameter server architectures and highlights some of the setbacks that they may face in practical settings.  Section~\ref{secRGCF} gives a detailed outline of the RGCF framework and the training procedure of the ANN-based filter.  Section~\ref{secResults} gives an analysis of the runtime of the RGCF framework to achieve Byzantine fault tolerance, shows our results on the resilience of the framework to an arbitrary number of workers, and the subsequent empirical communication overhead and gradient filtering overhead in comparison to the prior work.  Section~\ref{secConclusion} concludes and discusses future possibilities.

\section[Related Work]{Related Work}
\label{RecentWork}

The classical Byzantine Generals Problem~\cite{lamport1982byzantine} in distributed
computing describes a system where some of the agents or nodes (`Generals' in the story) are deceptive and secretly working against
the interests of the group as a whole, but are not known by the other
(non-Byzantine) agents to be deceptive.  In the classical formulation,
the problem is one of reaching \emph{consensus}, and it is known that
the number of Byzantine agents must be less than one-third of the
total number, for the group to be successful.

In the distributed computing literature~\cite{AW98}, a Byzantine
failure is an unrestricted and all-encompassing type that allows for
any and all failure modes (and intermittent failures), as opposed to
simple crash and such failures that allow only a single failure mode.

\begin{figure}
    \centering
    \includegraphics[scale=0.65]{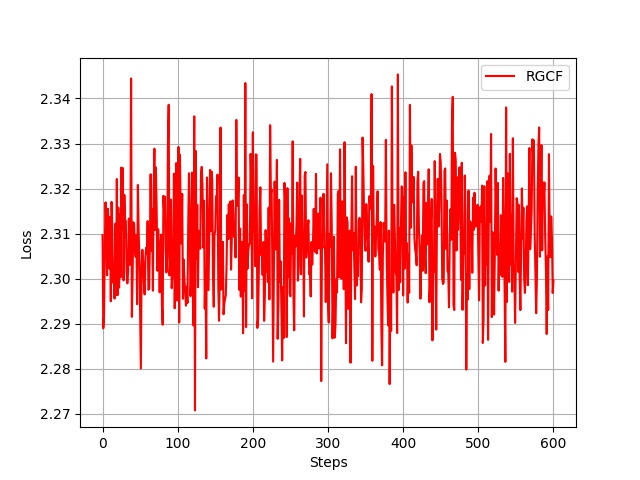}
    \caption{RGCF resilience towards 100\% Byzantine workers using the Inverse Attack}
    \label{fig:Hundred}
\end{figure}

This paper considers the standard parameter server architecture~\cite{li2014scaling} that consists of a single parameter server and $n$ workers.  Each worker samples $m$ (i.i.d) data points from distribution $D$ to form its local dataset $d$.  At each time step $t$, the parameter server picks a worker at random and passes the current model parameters $W_{t}$.  The worker samples a random mini-batch $b\sim d$ and computes its local gradient $\nabla_{W_{t}}(b)$ and passes it back to the central server. The central server then updates its parameters $W_{t}$ using the general gradient update $W_{t+1} = W_{t} - \alpha [\nabla_{W_{t}}(b)]$, where $\alpha$ is the preset learning learning rate of the parameter server.  Formally, we are trying to find the optimal parameters $W^*$ that minimizes the global loss function $L(W, x)$, 
\begin{align*}
   W^* = \arg \min_{W} L(W, x)
\end{align*}
 where $x$ is the local data sampled by the workers.
 
This problem has been extensively studied in the past; the algorithms from the literature~\cite{blanchard2017machine,allen2020byzantine,guerraoui2018hidden, chen2017distributed} provide theoretical guarantees for the convergence of the parameter server.  They consist mainly of two steps, in the first step they broadcast the model parameters $W_{t}$ to all worker nodes and each worker node $i$ computes its respective gradient $\nabla_{W}(b_{i})$ using a random mini-batch $b_{i}\sim d_{i}$ and passes it back to the parameter server.  In the second step, they use an aggregation function $Agg(\nabla_{W}(b_{1}), \nabla_{W}(b_{2}),\ldots, \nabla_{W}(b_{n}))$ that returns an estimate of the true gradient $\tilde{\nabla}_{W}$.  They then update the parameter server using the gradient update $W_{t+1} = W_{t} - \alpha [\tilde{\nabla}_{W}]$.  The aggregation functions can primarily be classified into two types: as those based on the geometric median \cite{blanchard2017machine, guerraoui2018hidden, chen2017distributed, allen2020byzantine}, and those based on the coordinate-wise median \cite{yin2018byzantine, yang2019byzantine, xie2018phocas, alistarh2018byzantine, xie2019slsgd}.

\begin{figure}
\begin{subfigure}{0.5\textwidth}
  \centering
  \includegraphics[scale=0.4]{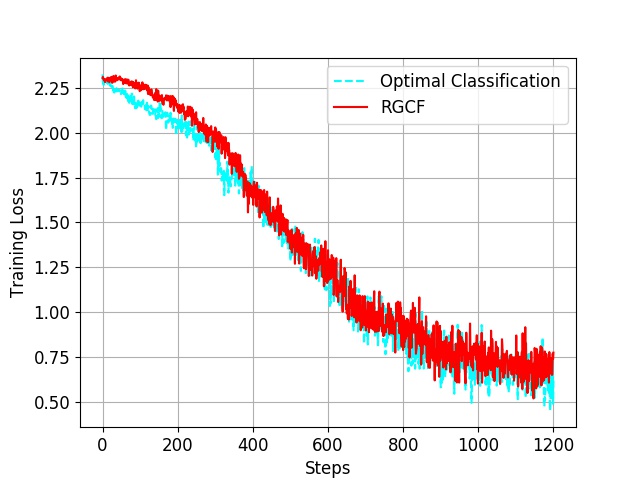}    
  \caption{}
  \label{fig:sub-first}
\end{subfigure}%
\hfill
\begin{subfigure}{0.5\textwidth}
  \centering
  \includegraphics[scale=0.4]{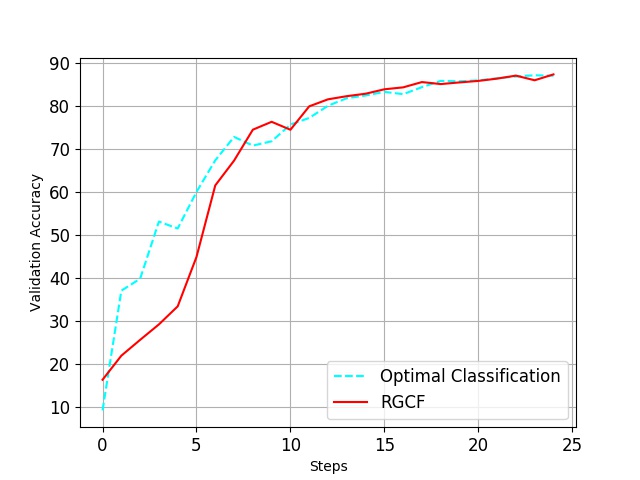}    
  \caption{}
  \label{fig:sub-second}
\end{subfigure}%
\caption{(a) Training loss and (b) Validation accuracy of the classification model when trained with 90\% Byzantine workers using the Inverse Scaled Gradient Attack}
\end{figure}

In the geometric median approach, it is assumed that the non-Byzantine gradients are clustered around the true gradient in $R^d$ space with distances measured using $l_p$ norms.  For this, they initially assume that a fraction $f$ of the gradients are Byzantine.  The aggregation function drops these $nf$ gradients (where $n$ is the number of workers) and estimates the true gradient as the geometric median of the remaining $n(1-f)$ gradients.  The foundational algorithm in this approach \textit{Krum} \cite{blanchard2017machine} defines a score function $S(v_{i}) = \sum_{i,j} || v_{i} - v_{j} ||$, where $j$ runs over the $n-f-2$ closest vectors to $v_i$.  The true gradient $v_{*}$ is estimated as $v_{*} = \arg \min_{v} S(v).$  In the Krum algorithm $f$ is a parameter that must be set before training.  Since $nf$ gradients are discarded in every round, they experience a slowdown in convergence proportional to $f$. In practice, if $f$ is larger than the actual fraction of Byzantine workers seen, there is an unnecessary slowdown as good gradients are also discarded.  On the other hand, if $f$ is smaller than the actual fraction of Byzantine workers, the convergence of the model suffers as even a single bad gradient can harm the convergence of the parameter server.  Krum also is limited by a hard bound on $f$, which must be lower than $1/2$ to guarantee the convergence of the parameter server; it also incurs a computational overhead of $\mathcal{O}(n^2d)$ per gradient update step.

Other notable works in this approach such as \cite{chen2017distributed} use the \textit{geometric median of means} to filter the Byzantine gradients, which can handle up to $\frac{n}{2}$ Byzantine workers and has a communication overhead of $\mathcal{O}(nd + nd\log^3(M))$, where $M$ is the size of the training dataset. \cite{allen2020byzantine} use the median of the sum of gradients generated by a worker over time to filter them as \textit{good} or \textit{bad}.  They remove the worker that generates \textit{bad} gradients from further considerations.  However, they also operate under the assumption that $f < \frac{n}{2}$.  It has been shown \cite{guerraoui2018hidden} that geometric median approaches suffer from the \textit{Curse of Dimensionality} as we move to higher-dimensional parameter spaces. This is because Byzantine workers can leverage the fact that it becomes hard to differentiate between gradients that differ slightly in all coordinates as compared to gradients that differ largely in a single coordinate as the $d$ increases.  This has been addressed by ~\cite{guerraoui2018hidden} who use the same aggregation function as Krum~\cite{blanchard2017machine} or Medoid to filter the bad gradients.  They then take the coordinate-wise median to obtain the true gradient. However, this method is bound by a threshold of $\frac{1}{4}$ for $f$. 

The coordinate median based approaches are similar~\cite{guerraoui2018hidden, yin2018byzantine, alistarh2018byzantine}, where coordinate medians are used in the aggregation function. \cite{yang2019byzantine} indicate that coordinate median approaches have a lower classification accuracy as compared to geometric median approaches. \cite{yang2019byzantine} have adopted a similar approach, where they use a hand-crafted solution based on the coordinate median.  They have a limit of $\frac{1}{2}$ on $f$, and incur a computational overhead of $\mathcal{O}(nd)$, which is better than the $\mathcal{O}(n^2d)$ that is encountered in the geometric median algorithms.

\begin{figure*}
\begin{subfigure}{0.45\textwidth}
  \centering
  \includegraphics[width=\textwidth]{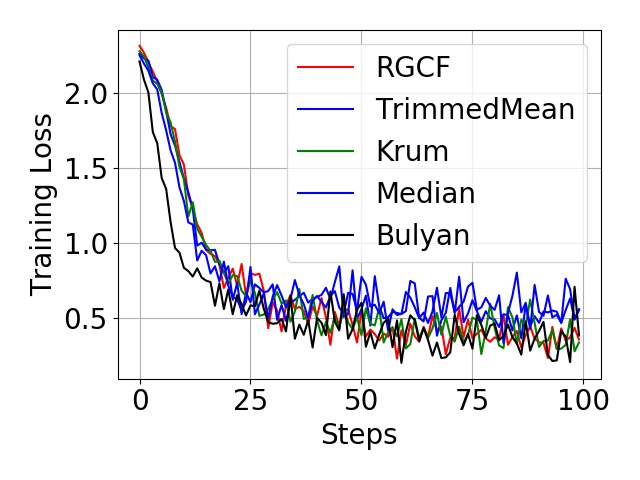}    
  \caption{}
  \label{fig:ones20}
\end{subfigure}%
\hfill
\begin{subfigure}{0.45\textwidth}
  \centering
  \includegraphics[width=\textwidth]{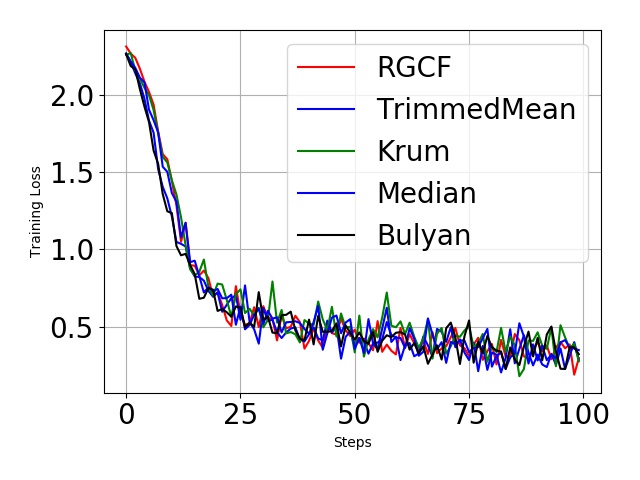}    
  \caption{}
  \label{fig:shift20}
\end{subfigure}%
\hfill
\begin{subfigure}{0.45\textwidth}
  \centering
  \includegraphics[width=\textwidth]{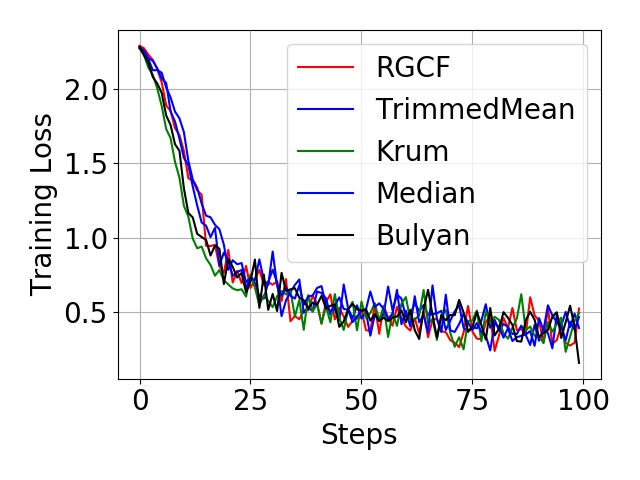}    
  \caption{}
  \label{fig:inv20}
\end{subfigure}%
\hfill
\begin{subfigure}{0.45\textwidth}
  \centering
  \includegraphics[width=\textwidth]{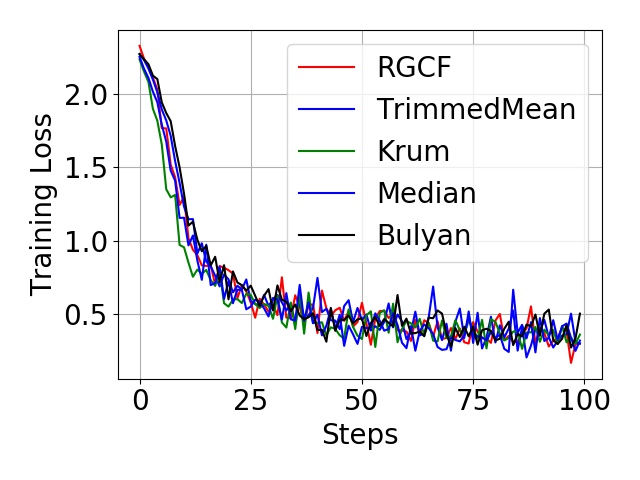}    
  \caption{}
  \label{fig:gauss20}
\end{subfigure}%
\caption{Comparative study of resilience of various methods when 20\% workers are using the (a) All Ones (b) Gradient Shift (c) Inverse (d) Random Gradient Attack.}
\label{l_20}
\end{figure*}

\begin{figure*}
\begin{subfigure}{0.45\textwidth}
  \centering
  \includegraphics[width=\textwidth]{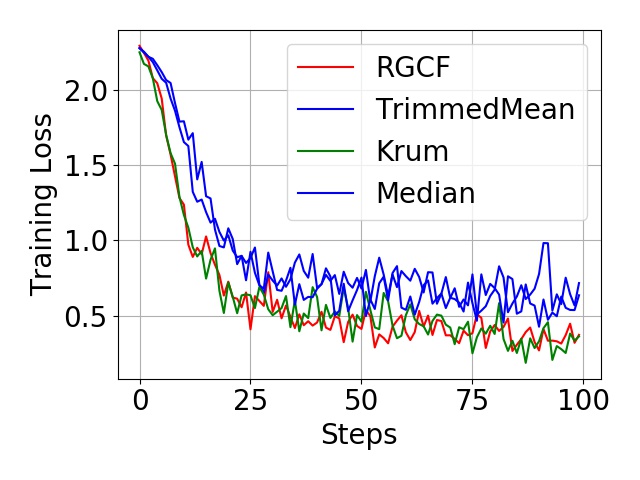}    
  \caption{}
  \label{fig:ones33}
\end{subfigure}%
\hfill
\begin{subfigure}{0.45\textwidth}
  \centering
  \includegraphics[width=\textwidth]{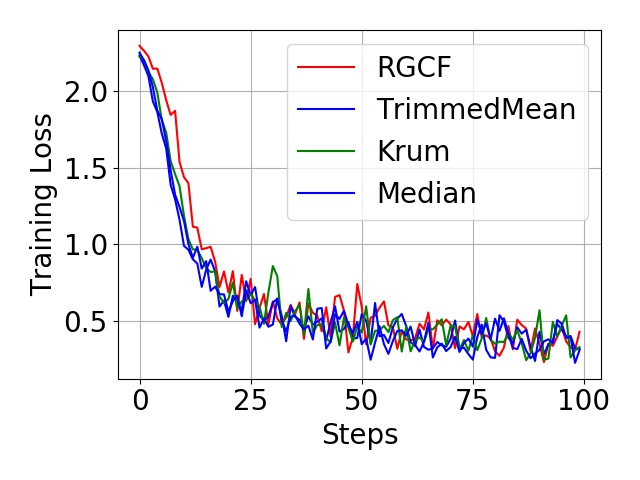}    
  \caption{}
  \label{fig:shift33}
\end{subfigure}%
\hfill
\begin{subfigure}{0.45\textwidth}
  \centering
  \includegraphics[width=\textwidth]{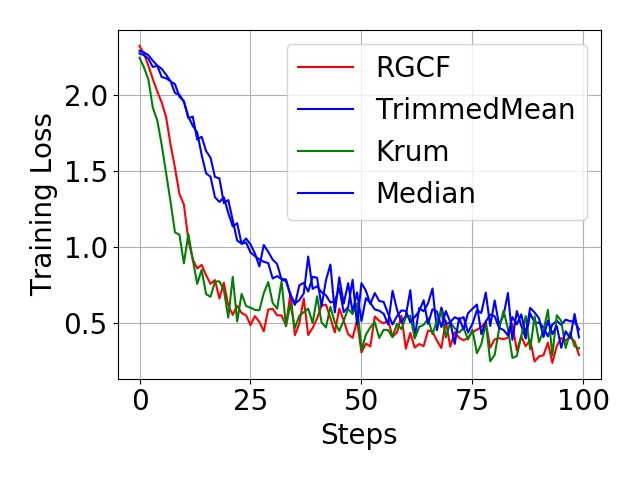}    
  \caption{}
  \label{fig:inv33}
\end{subfigure}%
\hfill
\begin{subfigure}{0.45\textwidth}
  \centering
  \includegraphics[width=\textwidth]{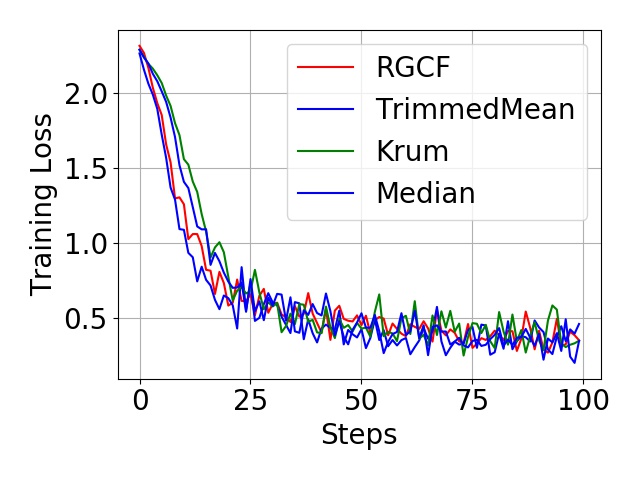}    
  \caption{}
  \label{fig:gauss33}
\end{subfigure}%
\caption{Comparative study of resilience of various methods when 33\% workers are using the (a) All Ones (b) Gradient Shift (c) Inverse (d) Random Gradient Attack.}
\label{l_33}
\end{figure*}

Other related works approach this problem differently.  ~\cite{cao2019distributed} is similar to our work to a certain extent, where they sample a noisy gradient from a clean local dataset and compare it with the received gradients to filter the Byzantine gradients.  They provide two algorithms: the first requires $f$ to be passed as a parameter, and the second does not.  They ensure convergence in the presence of an arbitrary number of Byzantine workers; however, they are currently limited to convex loss functions.  ~\cite{xie2018phocas} solve the generalized Byzantine problem where individual coordinates within a gradient can be Byzantine as opposed to the classical Byzantine problem that this paper solves, where entire gradients are classified as Byzantine.  ~\cite{xie2018zeno} are also similar to this work where they score gradients by the change in the loss function of the parameter server, and take the average of the $n-f$ gradients with the highest score; hence their performance also depends on the parameter $f$ that must be passed by the user.  ~\cite{damaskinos2018asynchronous} consider the case of asynchronous Stochastic Gradient Descent (SGD) in the presence of Byzantine workers and is not directly comparable with this work. 

~\cite{mahdi2020} consider the case where the parameter server itself can be Byzantine, and solve this by duplicating the parameter server multiple times and are able to guarantee convergence when $\frac{n}{3}$ servers and $\frac{n}{3}$ workers are Byzantine. They also solve this problem for the asynchronous SGD optimisation setting as well.  \cite{el2021collaborative} study the problem of \textit{Byzantine collaborative learning}, where they try to minimise a global loss function in a decentralised manner.  They remove the need for a parameter server and the workers learn via collaboration using one another's local data. ~\cite{damaskinos2019aggregathor} implement a framework over Tensorflow that implements some of their algorithms and assesses their practical running costs. The above solutions tackle the Byzantine fault tolerance problem for different frameworks than the one we are considering, and also require an $\frac{n}{3}$ bound on the number of workers that can be Byzantine.

The problem of Byzantine fault tolerance has been studied in other contexts apart from supervised learning.  For example, Chen et al.~\cite{chen2022} consider the problem of Byzantine agents in distributed RL training and give algorithms that facilitate online and offline learning in the presence of these Byzantine agents.

\section{Robust Gradient Classification Framework} \label{secRGCF}
In this section, we formally introduce the problem of Byzantine fault tolerance in the parameter server training setup and how it has been formulated using the RGCF framework. Next, we discuss the task of gradient classification using an artificial neural network and briefly indicate the motivation for the choice of this architecture.  After this, we give a detailed explanation of how the filter is trained via simulations, and the learning algorithm used to train the filter.

\subsection{Overview} \label{overviewSec}
This section explains our Robust Gradient Classification Framework (RGCF), which applies a modified parameter server architecture and includes a pattern recognition filter for classifying Byzantine gradients in distributed SGD.  The filter, when given a gradient $\nabla_{W}(x)$ and the scalar value of the loss $L(W, x)$, returns a scalar value $B\in\{0, 1\}$.  Each iteration $t$ of training the parameter server is split into five steps. 

In the first step, the parameter server picks a worker node based on some underlying strategy.  For simplicity we assume the strategy is picking a worker $i$ at random (line 5 in Algorithm \ref{RGCF}).

In the second step, the worker $i$ then samples a random mini-batch $b_{i}$ from its local dataset $d_{i}$ and receives parameters $W_{t}$ from the server.  The worker then computes the gradient $\nabla_{W_{t}}(b_{i})$ and returns it to the parameter server (line 6 in Algorithm \ref{RGCF}). Depending on the nature of the worker, it either returns the true gradient $\nabla_{W_{t}}(b_{i})$ or a Byzantine gradient $\neg\nabla_{W_{t}}(b_{i})$ (lines 1--6 of Worker in Algorithm \ref{RGCF}). 

During the third step, the parameter server passes the obtained gradient $\nabla_{W_{t}}(b_{i})$ as well as the scalar loss $L(W_{t},b_{i})$ to the filter which computes the value $B$. 

In the fourth step, the value $B$ is returned to the parameter server (line 7 in Algorithm \ref{RGCF}).

In the final step, the parameter server is then updated by the equation:
\begin{equation} \label{gradUpdateEquation}
    W_{t+1} = W_{t} - (1-B) \alpha \nabla_{W_{t}}
\end{equation}

In Figure \ref{architectureFig} the workflow can be explained as (refer to the numbers assigned to the arrows):
\begin{enumerate}
    \item A random worker receives the model parameters from the parameter server.
    \item The worker computes its gradient $\nabla_{W_{t}}(b_{i})$  and sends the same to the server.
    \item The server sends this gradient $\nabla_{W_{t}}(b_{i})$ to the filter.
    \item The filter returns a single scalar value $B$ denoting whether to accept or reject the gradient.
    \item At inference time the server then updates its parameters using the belief $B$ to according to equation \eqref{gradUpdateEquation}. When training the filter via simulation, the parameter $B$ is evaluated against the ground truth to give feedback to the filter. (Lines 13--14 Algorithm \ref{RGCF})
    
\end{enumerate}

\begin{algorithm}[H]
    \caption{Robust Gradient Classification Framework}
    \label{RGCF}

{\textbf{Server}}
\begin{algorithmic}[1]
        \STATE $T \leftarrow$ number of steps per episode.
        \STATE $W \leftarrow$ parameters of the server.
         \STATE $\alpha \leftarrow$ learning rate of parameter server.
        \FOR{$k = 0,1,2,...,T$}
        \STATE $i \xleftarrow[]{}$ sample from list of workers.
        \STATE $\nabla_{W} \xleftarrow[]{}$  \text{Worker}(i, $W_{t}$) \COMMENT{returns generated gradient}
        \STATE $B  \xleftarrow[]{}$ \text{Filter (}$\nabla_{W}$, $L$, $\beta$, $r$) 
        \STATE $W_{t+1} \xleftarrow[]{} W_{t} - (1 - B) \alpha \nabla_{W}$
        \ENDFOR
\end{algorithmic}
{\textbf{Worker($id$, $W_{t}$)}}
\begin{algorithmic}[1]
        \STATE $b_{i} \xleftarrow[]{}$  \text{sample from local dataset }$d_{i}$.
        \IF{$id$ is Byzantine}
        \STATE \textbf{return} $\neg\nabla_{W}$($b_{i}$)
        \ELSE
        \STATE \textbf{return} $\nabla_{W}$($b_{i}$)
        \ENDIF
\end{algorithmic}
\end{algorithm}

\subsection{Classifying Byzantine Gradients}
We use pattern recognition models to classify Byzantine gradients.  To classify a gradient, we require two key features, the first being the direction of the gradient, and the other being the location of the model parameters on the loss manifold.  We thus provide the filter an input of the gradient $\nabla_{W_{t}}(b_{i})$, and as an approximation of the location of the model parameters, we provide the filter with a scalar value of the current loss $L(W_{t},b_{i})$. 

We experimented with a number of gradient-based classification models such as logistic regression, Linear SVM, and artificial neural networks (ANN). The models that are most suitable are the ones that can be trained via online learning due to the nature of the episodic training of the filter, explained in the following subsection.  Simple linear models such as logistic regression and linear SVM were not able to classify the gradients accurately.  Kernel SVMs cannot be trained in an online manner via SGD, hence we used neural networks to classify Byzantine gradients.

\subsection{Training the Filter}
\label{TrainingFilter}

The filter is trained via simulation of the training procedure of the parameter server on a local dataset $D_{local}$.  The simulation consists of $N$ episodes. In each episode we train the parameter server model from scratch using $D_{local}$. This training scenario consists of 1 Byzantine worker and 1 honest worker.  This ensures an equal distribution of true and Byzantine gradients encountered by the parameter server in each episode.  At each step in the simulation, a worker $i$ is randomly sampled by the parameter server with equal probability (line 8 in Algorithm \ref{trainingsimulation}).  The worker then returns the true gradient if it is an honest worker, and a Byzantine gradient otherwise (line 9 in Algorithm \ref{trainingsimulation}).  In this simulation, the Byzantine worker returns a gradient where each coordinate is sampled from a normal distribution with mean 0 and variance 1, which is also termed as the Random Gradient Attack.  We can see from Section~\ref{secResults} that training the filter to defend against the Random Gradient Attack generalizes its resistance to a wide range of attacks.

We refer to a data point as the computed gradient $\nabla_{W_{t}}(b_{i})$ appended with the scalar loss $L(W_{t},b_{i})$.  The filter then predicts the class of the data point, where class 1 implies that a gradient is Byzantine, and class 0 otherwise (line 11 in Algorithm \ref{trainingsimulation}).  The filters' weights are updated using weighted binary cross entropy loss between the predicted class of the gradient and the ground truth (lines 13--14 in Algorithm \ref{trainingsimulation}).  Class 1 is given a weight of 10.  Thus, the filter is trained in an online manner via stochastic gradient descent using the incoming gradients one at a time from the simulation.  We train the filter in this manner because the size of the gradients can be large, hence storing the computed gradients over an entire episode and then training the filter offline would require a large amount of memory.  Note that it is also possible to take an in-between approach to store small batches of computed gradients in a buffer, and then train the filter on the data points in the buffer, before clearing it to store new data points. 

For efficient training, the simulation consists of optimal training episodes of the parameter server---that is, all Byzantine gradients are dropped and all true gradients are passed to the parameter server (line 12 in Algorithm \ref{trainingsimulation}).  In the update step in the simulation (Algorithm \ref{TrainingFilter}, line 12), the true value of the nature of the worker $B$ is used and not the predicted value of the filter $\hat{B}$. This is done because using the predicted values of the untrained filter could disrupt convergence of the parameter server in the simulation, requiring more episodes for the filter to be trained. 

\begin{algorithm}[H]
    \caption{Training the Filter}
    \label{trainingsimulation}
{\textbf{Server}}
\begin{algorithmic}[1]
        \STATE $N \leftarrow$ number of training episodes.
        \STATE $T \leftarrow$ number of steps per episode.
        \STATE $\theta \leftarrow$ parameters of the filter.
        \STATE $\alpha_1, \alpha_2 \leftarrow$ learning rate of parameter server, filter. 
        \STATE $p \leftarrow$ positive weight.
        \FOR {$n \leftarrow \{1, \ldots, N\}$}
        \STATE $W \sim$ Random Init 
        \FOR {$t \leftarrow \{1, \ldots, T\}$}
        {\STATE $i \xleftarrow[]{}$ sample from list of workers.
        \STATE $B \leftarrow 1$ if $i$ is Byzantine, else $0$.  
        \STATE $\nabla_{W} \xleftarrow[]{}$  \text{Worker}(i, $W_{t}$)
        \STATE $\hat{B}  \xleftarrow[]{}$ \text{Filter (}$\nabla_{W}$, $L$, $\beta$, $r$) 
        \STATE $W_{t+1} \xleftarrow[]{} W_{t} - (1 - B) \alpha_1 \nabla_{W}$}
        \STATE $L_{\theta} = p[\mathbb{1}(B=1)\log(\hat{B})] + \mathbb{1}(B=0)\log(1-\hat{B})$.
        \STATE $\theta \leftarrow \theta - \alpha_2 \nabla_{L_{\theta}}$
        
        \ENDFOR
        \ENDFOR
\end{algorithmic}
\end{algorithm}

\section{Experiments and Results} \label{secResults}

In this section, we first describe our
training setup, which includes our distributed SGD framework to train an image classifier, and the filter. Subsequently we demonstrate the robustness of our framework towards an arbitrary number of Byzantine workers, and compare our performance against various attacks with the prior work.  Finally we analyse the runtime of our solution and empirically verify that it achieves a $10$x speedup over the prior work.

\begin{table*}[h]
    \begin{subtable}[h]{0.45\textwidth}
        \centering
        \begin{tabular}{ p{1cm} p{0.8cm} p{0.8cm} p{0.8cm} p{0.8cm}}
 \toprule
 \multicolumn{5}{c}{Inverse Attack} \\
 \midrule

 $f$ & 20\% & 33\% & 50\% & 90\%\\
 \midrule
 RGCF  & \cmark   &  \cmark &   \cmark & \cmark \\
 Krum &  \cmark  & \cmark  & \xmark & \xmark\\
 Bulyan & \cmark & \textendash & \textendash &\textendash\\
 T-Mean  & \cmark & \cmark &  \xmark & \xmark\\
 Median & \cmark & \cmark & \xmark  & \xmark\\
 \bottomrule
 
\end{tabular}
\caption{}
 \label{inverseTable}
    \end{subtable}
    \hfill
    \begin{subtable}[h]{0.45\textwidth}
        \centering
        \begin{tabular}{ p{1cm} p{0.8cm} p{0.8cm} p{0.8cm} p{0.8cm}  }
 \toprule
 \multicolumn{5}{c}{Random Gaussian Attack} \\
 \midrule
 $f$ & 20\% &  33\% & 50\% & 90\%\\
 \midrule
 RGCF  & \cmark   &  \cmark &   \cmark & \cmark \\
 Krum &  \cmark  & \cmark  & \cmark & \xmark\\
 Bulyan & \cmark & \textendash & \textendash &\textendash\\
 T-Mean  & \cmark & \cmark &  \xmark & \xmark\\
 Median & \cmark & \cmark & \cmark  & \xmark\\
 \bottomrule
\end{tabular}
\caption{}
\label{randomTable}
     \end{subtable}
     
     \vspace{0.5cm}
     
     \begin{subtable}[h]{0.45\textwidth}
        \centering
        \begin{tabular}{ p{1cm} p{0.8cm} p{0.8cm} p{0.8cm} p{0.8cm}  }

 \toprule
 \multicolumn{5}{c}{Gradient Shift Attack} \\
 \midrule
 $f$ & 20\% &  33\% & 50\% & 90\%\\
 \midrule
 RGCF  & \cmark   &  \cmark &   \cmark & \cmark \\
 Krum &  \cmark  & \cmark  & \cmark & \xmark\\
 Bulyan & \cmark & \textendash & \textendash &\textendash\\
 T-Mean  & \cmark & \cmark &  \xmark & \xmark\\
 Median & \cmark & \cmark & \xmark  & \xmark\\
 \bottomrule
\end{tabular}
\caption{}
\label{shiftTable}
    \end{subtable}
    \hfill
    \begin{subtable}[h]{0.45\textwidth}
        \centering
        \begin{tabular}{ p{1cm} p{0.8cm} p{0.8cm} p{0.8cm} p{0.8cm}  }

 \toprule
 \multicolumn{5}{c}{All Ones Attack} \\
 \midrule
 $f$ & 20\% &  33\% & 50\% & 90\%\\
 \midrule
 RGCF  & \cmark   &  \cmark &   \cmark & \cmark \\
 Krum &  \cmark  & \cmark  & \xmark & \xmark\\
 Bulyan & \cmark & \textendash & \textendash &\textendash\\
 T-Mean  & \cmark & \cmark &  \xmark & \xmark\\
 Median & \cmark & \cmark & \xmark  & \xmark\\
 \bottomrule
 
\end{tabular}
\caption{}
\label{onesTable}
     \end{subtable}
     \caption{Resilience of the algorithms against different attack types for various values of $f$}
     \label{tab:temps}
     
\end{table*}

\subsection{Training Details}
To train and validate our framework, we looked at the task of image classification in an online distributed setting in the presence of Byzantine workers.  For this, we used a medium-size neural network which comprises of two convolutional layers with kernel size being 3$\times$3, followed by a max pooling layer of stride 2.  The generated image features are passed through two fully-connected layers, and finally a sigmoid layer.  This neural network was implemented using the PyTorch framework \cite{NEURIPS2019_9015}.  The CNN was trained on the MNIST~\cite{lecun1998gradient} dataset.  The CNN was trained with a learning rate of 0.01. with a batch size of 128, and was trained using standard distributed SGD.

The filter is an artificial neural network that consists of 2 hidden layers with each consisting of $64$ and $32$ neurons respectively.  The filter was trained using just one training simulation of 500 steps, with 10,000 images that constitutes the workers' local dataset.  As stated in the Training the Filter subsection, the simulation consists of one Byzantine agent and one honest worker, and the Byzantine agent is assumed to employ the Random Scaled Gradient Attack.  The filter was trained using the Adam optimiser~\cite{kingma2014adam}.  The framework was then validated over a training instance of the classification model from scratch using 50,000 images.  The framework was then evaluated with $20\%, 33\%, 50\%, 90\%$ and $100\%$ of the workers being Byzantine respectively. 

We test our framework against five types of attacks from Byzantine workers, as seen in prior work~\cite{data2021byzantine}.  (Other works~\cite{blanchard2017machine, guerraoui2018hidden} use a subset of these attacks.) 
\begin{enumerate}
    \item Random Scaled Gradient Attack: \\
    The gradient is a scaled random Gaussian vector.
    \item Inverse Scaled Gradient Attack: \\
    A scaled version of the true gradient with its direction reversed.
    \item All Ones: \\
   The gradient contains all ones.
    \item Gradient Shift Attack: \\
    The gradient is shifted by a scaled (50) random Gaussian vector.
\end{enumerate}

We evaluate our framework against Krum \cite{blanchard2017machine}, Coordinate-Wise Trimmed Mean \cite{xie2018zeno, xie2019slsgd},  Coordinate Median \cite{yin2018byzantine}, and Bulyan \cite{guerraoui2018hidden} (refer to Related Work for a brief overview of these methods).

We show the robustness and efficiency of our framework based on two metrics: robustness to various fractions of Byzantine workers, and runtime in practice.

\begin{figure*}
\begin{subfigure}{0.45\textwidth}
  \centering
  \includegraphics[width=\textwidth]{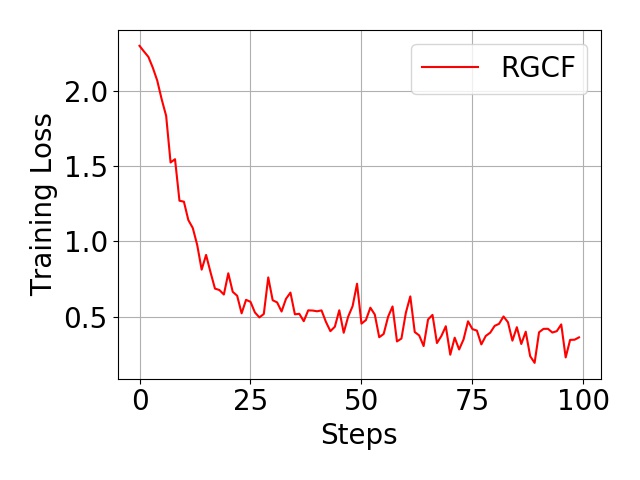}    
  \caption{}
  \label{fig:ones50}
\end{subfigure}%
\hfill
\begin{subfigure}{0.45\textwidth}
  \centering
  \includegraphics[width=\textwidth]{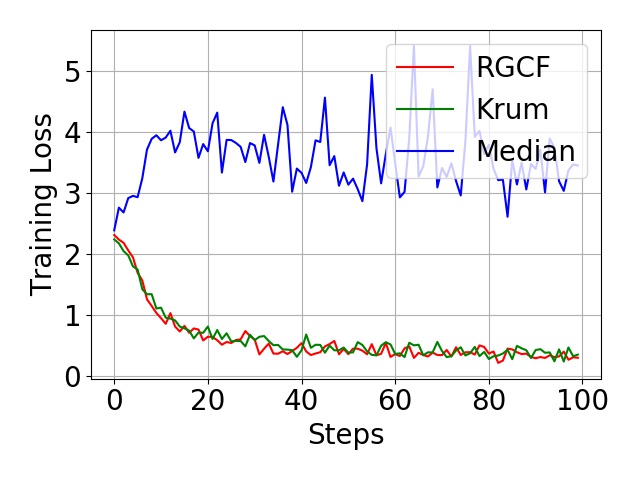}    
  \caption{}
  \label{fig:shift50}
\end{subfigure}%
\hfill
\begin{subfigure}{0.45\textwidth}
  \centering
  \includegraphics[width=\textwidth]{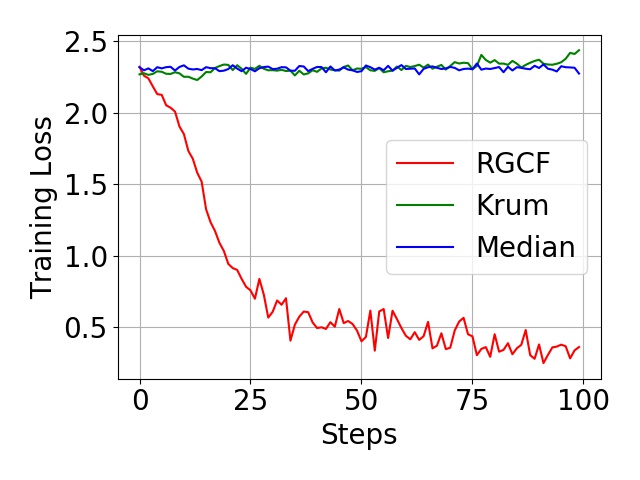}    
  \caption{}
  \label{fig:inv50}
\end{subfigure}%
\hfill
\begin{subfigure}{0.45\textwidth}
  \centering
  \includegraphics[width=\textwidth]{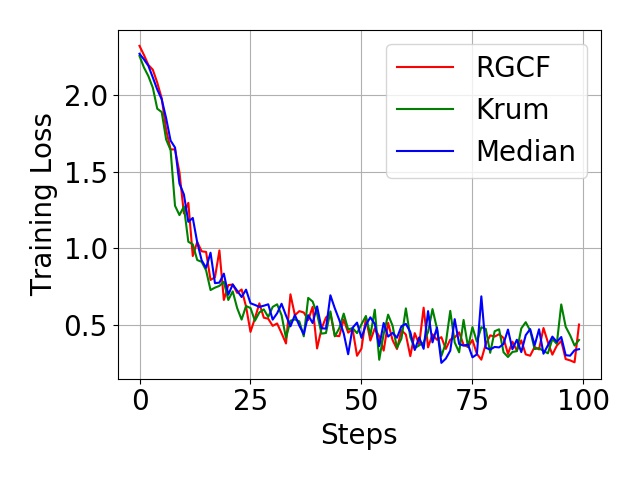}    
  \caption{}
  \label{fig:gauss50}
\end{subfigure}%
\caption{Comparative study of resilience of various methods when 50\% workers are using the (a) All Ones (b) Gradient Shift (c) Inverse (d) Random Gradient Attack.}
\label{l_50}
\end{figure*}
\begin{figure*}
\begin{subfigure}{0.45\textwidth}
  \centering
  \includegraphics[width=\textwidth]{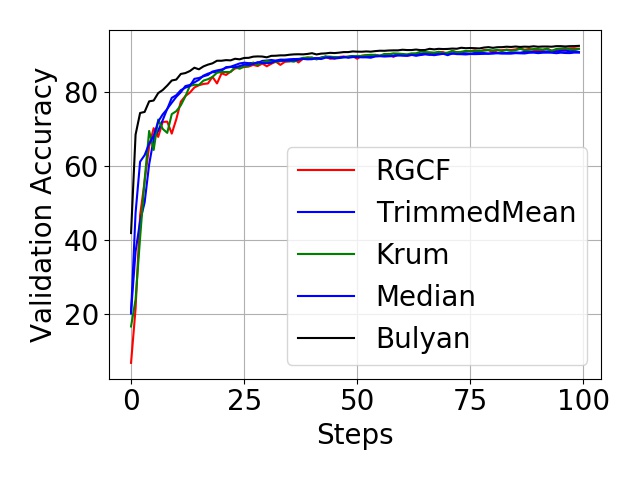}    
  \caption{}
  \label{fig:v_ones20}
\end{subfigure}%
\hfill
\begin{subfigure}{0.45\textwidth}
  \centering
  \includegraphics[width=\textwidth]{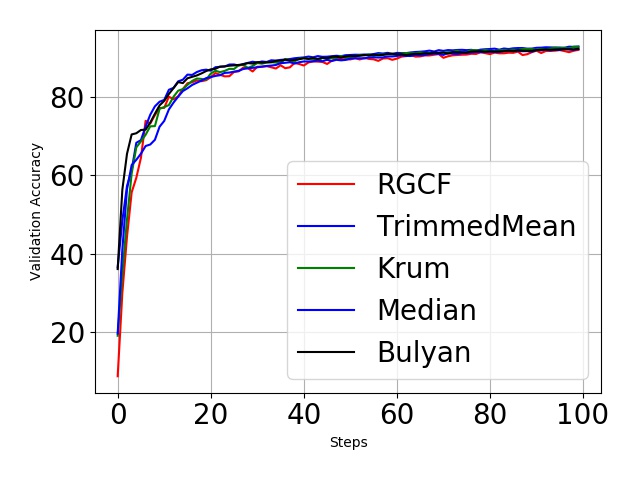}    
  \caption{}
  \label{fig:v_shift20}
\end{subfigure}%
\hfill
\begin{subfigure}{0.45\textwidth}
  \centering
  \includegraphics[width=\textwidth]{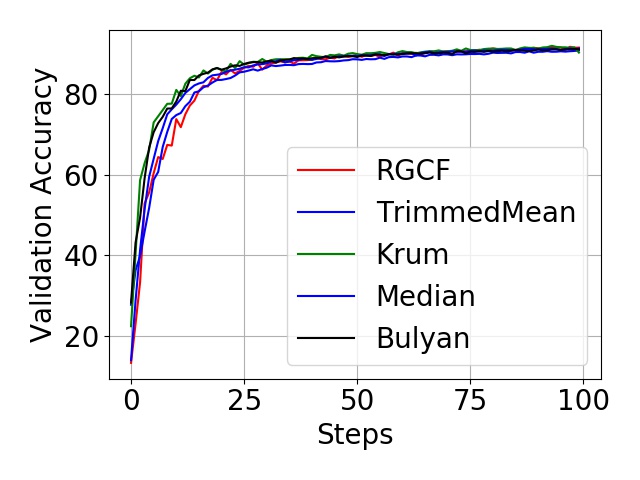}    
  \caption{}
  \label{fig:v_inv20}
\end{subfigure}%
\hfill
\begin{subfigure}{0.45\textwidth}
  \centering
  \includegraphics[width=\textwidth]{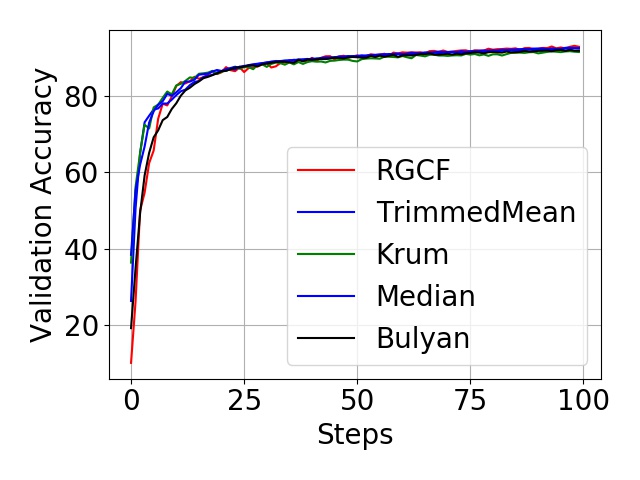}    
  \caption{}
  \label{fig:v_gauss20}
\end{subfigure}%
\caption{Comparative study of resilience of various methods when 20\% workers are using the (a) All Ones (b) Gradient Shift (c) Inverse (d) Random Gradient Attack.}
\label{v_20}
\end{figure*}
\begin{figure*}
\begin{subfigure}{0.45\textwidth}
  \centering
  \includegraphics[width=\textwidth]{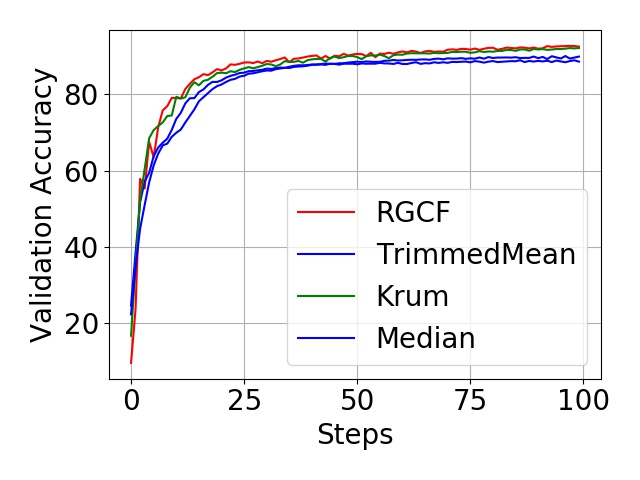}    
  \caption{}
  \label{fig:v_ones33}
\end{subfigure}%
\hfill
\begin{subfigure}{0.45\textwidth}
  \centering
  \includegraphics[width=\textwidth]{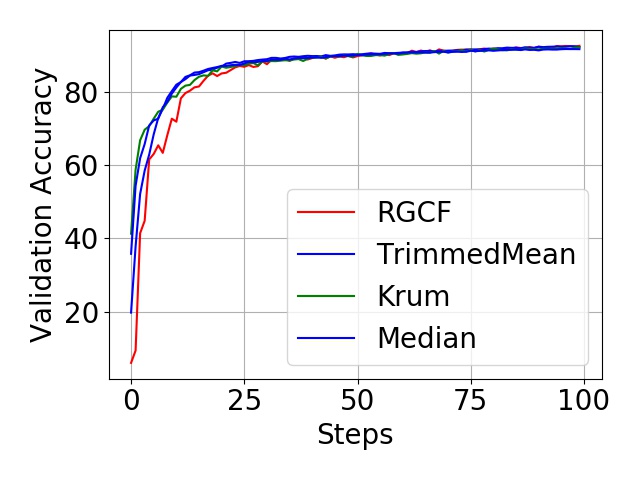}    
  \caption{}
  \label{fig:v_shift33}
\end{subfigure}%
\hfill
\begin{subfigure}{0.45\textwidth}
  \centering
  \includegraphics[width=\textwidth]{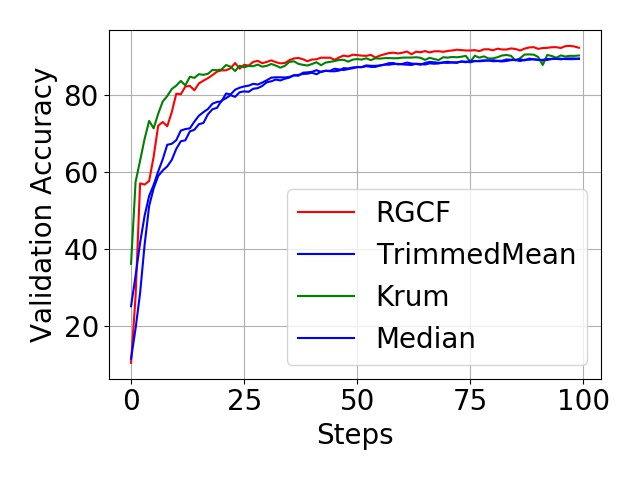}    
  \caption{}
  \label{fig:v_inv33}
\end{subfigure}%
\hfill
\begin{subfigure}{0.45\textwidth}
  \centering
  \includegraphics[width=\textwidth]{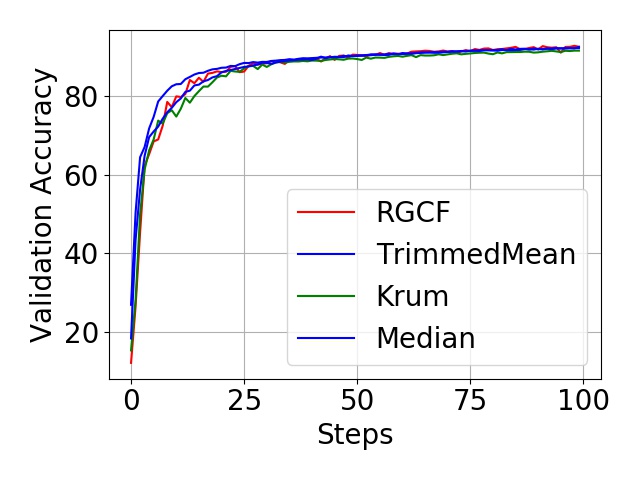}    
  \caption{}
  \label{fig:v_gauss33}
\end{subfigure}%
\caption{Comparative study of resilience of various methods when 33\% workers are using the (a) All Ones (b) Gradient Shift (c) Inverse (d) Random Gradient Attack.}
\label{v_33}
\end{figure*}
\begin{figure*}
\begin{subfigure}{0.45\textwidth}
  \centering
  \includegraphics[width=\textwidth]{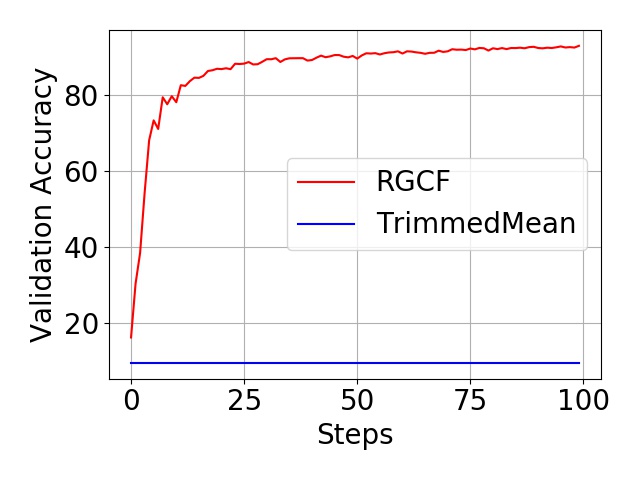}    
  \caption{}
  \label{fig:v_ones50}
\end{subfigure}%
\hfill
\begin{subfigure}{0.45\textwidth}
  \centering
  \includegraphics[width=\textwidth]{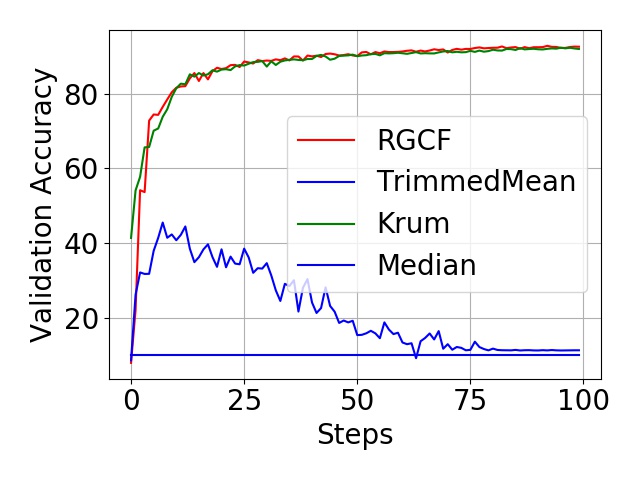}    
  \caption{}
  \label{fig:v_shift50}
\end{subfigure}%
\hfill
\begin{subfigure}{0.45\textwidth}
  \centering
  \includegraphics[width=\textwidth]{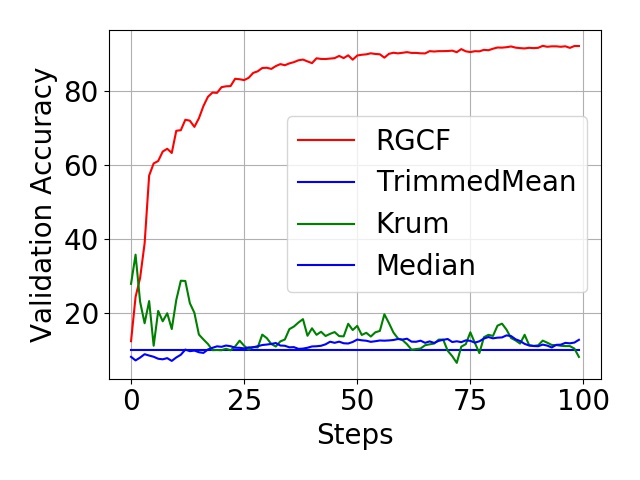}    
  \caption{}
  \label{fig:v_inv50}
\end{subfigure}%
\hfill
\begin{subfigure}{0.45\textwidth}
  \centering
  \includegraphics[width=\textwidth]{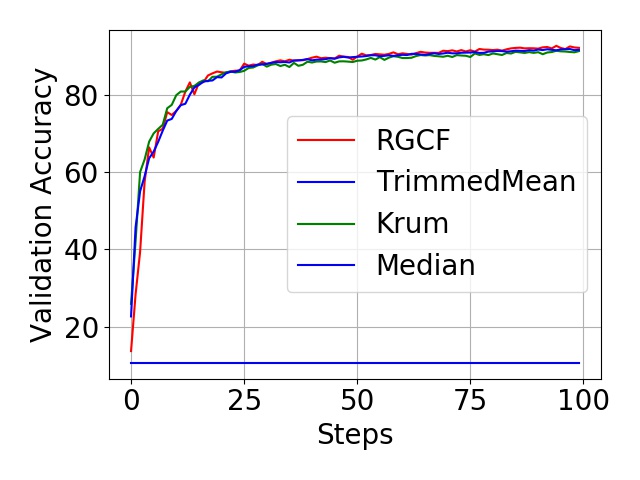}    
  \caption{}
  \label{fig:v_gauss50}
\end{subfigure}%
\caption{Comparative study of resilience
e of various methods when 50\% workers are using the (a) All Ones (b) Gradient Shift (c) Inverse (d) Random Gradient Attack.}
\label{v_50}
\end{figure*}

\subsection{Robustness to Arbitrary Numbers of Workers}
In the first experiment, we compare the different algorithms against a varying fraction of Byzantine workers.  We test the robustness against three values of $f: 20\%, 33\%$ and $50\%$. 

In Tables \ref{inverseTable}, \ref{randomTable}, \ref{shiftTable}, \ref{onesTable}
and Figures \ref{l_20}, \ref{l_33}, \ref{l_50}, \ref{v_20}, \ref{v_33} and \ref{v_50} we provide a summary of the performance of various algorithms against different types of adversarial attacks. A checkmark indicates that the parameter server is able to converge using the given algorithm in the presence of $f$ Byzantine workers.
We can observe that RGCF is able to converge in all cases.  Note that Bulyan operates only when $n \geq 4f + 3$, as a result we show the result only for the case of $f = 20\%$.

In the Inverse Attack and the All Ones Attack (Table \ref{inverseTable}, \ref{onesTable}) none of the other algorithms are able to achieve convergence of the parameter server when $f \geq 50$.

In the case of the Gradient Shift and Random Gaussian Attacks (Table \ref{randomTable}, \ref{shiftTable}), Krum is able to converge at $f = 50\%$ while the other algorithms fail to do so.

Figure \ref{fig:inv50} is one example of the loss of the classification model when 50\% of the agents are Byzantine and the attack employed is the Inverse Gradient Attack.  In this case we can see that RGCF achieves convergence of the parameter server and other algorithms fail.  We did not include the trimmed mean algorithm as the loss grows very rapidly to an extremely large value. We also did not include the Bulyan algorithm as it does not work for such a high value of $f$.  We obtain similar results for the other attack types as well.

Figures \ref{fig:sub-first} and \ref{fig:sub-second} compare the performance of our trained ANN-based filter against an optimal filter, i.e, a filter that drops every Byzantine gradient and accepts every true gradient.  We can see that even when $90\%$ of the workers are Byzantine, our filter performs close to optimal.

Next, we test the robustness of the filter when 100\% of the workers are Byzantine. The RGCF framework manages to reject all the gradients even when every worker is Byzantine.  We can see this from Figure \ref{fig:Hundred}. Note that the zigzag nature of figure \ref{fig:Hundred} is due to the stochastic nature of the mini-batches and not due to a change in the model parameters.

\subsection{Comparison of Runtimes} 
The runtime of any method depends on two factors: (a) the communication overhead, i.e., the number of workers that have to be queried to generate a gradient in one training step; and (b) the runtime of the gradient filtering step. We explain each factor in detail below.

\underline{Communication Overhead}:
Our framework requires just one gradient per step, while the other methods require a gradient from the all the workers. Thus, our method has an $\mathcal{O}(d)$ communication overhead, as compared to $\mathcal{O}(d \times n)$ for the others.

\underline{Gradient Filtering Overhead}:
The asymptotic runtime of our algorithm is the inference time of the neural network---a sequence of matrix multiplications.  Hence, the asymptotic runtime is $\mathcal{O}(\sum (l_{i}\cdot l_{i+1}))$, where $l_{i}$ is the number of neurons in the layer $i$.  $d$ is the size of the input layer or $l_{0}$.  In our case $d >> l_{i} \text{ for } i \neq 0$, so our asymptotic runtime is $\mathcal{O}(d \times h)$ where $h = l_{1}$. 
In practice, the empirical inference time of the neural network is quite low, due to efficient parallel implementations of matrix multiplications on GPUs.

In Table \ref{runtimeTable}, we compare the empirical runtime of one gradient update step of the parameter server, i.e., the communication overhead plus the gradient filtering overhead of RGCF, with the prior work.  These results are obtained by averaging over 100 gradient update steps with 10 workers.  We can see that RGCF is approximately $10$ times faster than the other algorithms.  It is also interesting to note that the communication overhead as well as the gradient filtering overhead of RGCF does not depend on the number of workers, hence RGCF can be used in training instances with a large number of workers without any loss in performance. 

\begin{table}
\centering
\caption{Comparison of empirical and asymptotic runtimes of different algorithms}
\label{runtimeTable}
\begin{tabular}{ p{2.5cm} p{2.5cm} p{2.5cm}}%

 \toprule
 \multicolumn{3}{c}{Runtimes} \\
 \midrule
 Algorithm & Empirical & Asymptotic \\
 \midrule
 RGCF  & 0.063 $\pm$ 0.013  & $\mathcal{O}(d \cdot h)$\\
 Krum &  0.520 $\pm$ 0.007 & $\mathcal{O}(d\cdot n^2)$\\
 Bulyan & \hspace{-0.054in}42.591 $\pm$ 0.950 & $\mathcal{O}(d \cdot n^2)$\\
 T-Mean  & 0.511 $\pm$ 0.005 & $\mathcal{O}(d\cdot n\log{n})$\\
 Median & 0.507 $\pm$ 0.006 & $\mathcal{O}(d \cdot n\log{n} )$\\
 \bottomrule
\end{tabular}
\end{table}

\section{Conclusion} \label{secConclusion}

With the increase in data being generated from multiple devices, there
has been a shift in methodology in how machine learning is carried
out.  In the classical approach, the entire dataset was stored and the
model was trained on a single machine.  Such models are also typically
trained on data that can be manually curated and labeled.  However, in
a distributed setting where live data comes from multiple
sources~\cite{gomes2019}, the model is prone to erroneous learning from
Byzantine gradients sampled from biased sources.  The distributed
setting also introduces the possibility of communication and
computational errors which can corrupt the gradients sent by the
workers.

Byzantine fault tolerance in distributed systems is well studied;
there have been numerous algorithms based on consensus proposed over
the years to tackle this problem.  Due to the above issues, there is a
need to extend such concepts to classical machine learning algorithms
such as SGD in distributed learning.

Distributed machine learning~\cite{verbraeken2021} is thus of much 
contemporary interest, and several solutions have been proposed for the 
problem of `Byzantine' machine learning~\cite{mahdi2020,blanchard2017machine,guerraoui2018hidden}.  

This paper describes RGCF, a method for achieving Byzantine fault
tolerance in distributed SGD for an arbitrary number of Byzantine
workers.  RGCF does not require an assumption of the number of
Byzantine workers, and has a communication overhead of
$\mathcal{O}(d)$ for each gradient update step as compared to
$\mathcal{O}(nd)$ encountered in the prior work.  It has a
running time of $\mathcal{O}(d\cdot h)$, where $d$ is the dimension of the gradient and $h$ is the number of neurons in the first hidden layer.  The performance of RGCF does
not depend on the number of workers $n$ and can scale effectively to a
large number of workers. 

In general, RGCF is suitable in cases where one has access to a local clean dataset and the parameter server has to be trained in an online manner. That is, the workers perpetually gain access to more and more data, and hence, the parameter server keeps improving on the specified task. 

For instance, RGCF can be used in the case where a language model is used for autocomplete on a user device \cite{autocomplete}. The language model is stored at a central server and the workers are devices such as laptops and mobile phones where autocomplete is used.  The central server's dataset to train the RGCF filter can be created by querying a small control group of workers that we know are honest; these workers however cannot generate enough data to train the large language model to convergence. After training the RGCF filter, the server is free to obtain gradients from an arbitrary number of workers.  RGCF can be extended in the future to the case of asynchronous
stochastic gradient descent where the server does not need to wait to receive the gradients from the workers which leads to faster training time.

\end{document}